\documentclass[runningheads]{llncs}
\usepackage{graphicx}
\usepackage{color}
\usepackage{multirow}
\usepackage{hhline}
\usepackage{latexsym}
\usepackage{stmaryrd}
\usepackage{hyperref}

\begin{document}

\title{How to find a good image-text embedding\\ for remote sensing visual question answering?}
\titlerunning{Which image-text embedding for remote sensing VQA?}
\author{Christel Chappuis\inst{1}\orcidID{0000-0002-9622-5428} \and
Sylvain Lobry\inst{2}\orcidID{0000-0003-4738-2416} \and
Benjamin Kellenberger\inst{1}\orcidID{0000-0002-2902-2014} \and
Bertrand Le Saux\inst{3}\orcidID{0000-0001-7162-6746} \and
Devis Tuia\inst{1}\orcidID{0000-0003-0374-2459}}
\authorrunning{C. Chappuis et al.}
\institute{Environmental Computational Science and Earth Observation Laboratory (ECEO), EPFL, Sion, Switzerland \and
Laboratoire d'Informatique Paris Descartes (LIPADE), Université de Paris, France \and
$\mathrm\Phi$-lab, European Space Research Institute (ESRIN), European Space Agency (ESA), Frascati, Italy}
\maketitle %
\begin{abstract}
Visual question answering (VQA) has recently been introduced to remote sensing to make information extraction from overhead imagery more accessible to everyone. VQA considers a question (in natural language, therefore easy to formulate) about an image and aims at providing an answer through a model based on computer vision and natural language processing methods. As such, a VQA model needs to jointly consider visual and textual features, which is frequently done through a fusion step. In this work, we study three different fusion methodologies in the context of VQA for remote sensing and analyse the gains in accuracy with respect to the model complexity. Our findings indicate that more complex fusion mechanisms yield an improved performance, yet that seeking a trade-off between model complexity and performance is worthwhile in practice.

\end{abstract}

\section{Introduction}
Over the last decades, large quantities of information about our planet have been recorded, in particular through Earth observation imagery. With this comes the need to develop appropriate tools to extract useful knowledge, especially in regard to environmental monitoring. However, developing such tools requires a strong technical knowledge (in image processing, machine learning, \emph{etc.}), which can limit the use of remote sensing imagery for applications with impact on daily life, such as urban planning, agriculture and environment.

Recently, a new task to access image information in human language emerged, known as visual question answering (VQA~\cite{antol_vqa_2015}): here, a model aims at providing an answer to an open-ended question in natural language about an image. Considering the wide variety of images and potential inquiries, the task of VQA is as challenging as appealing. For remote sensing VQA in particular, research is motivated by the prospect of opening the use of remote sensing imagery, making it available as a tool for anyone to benefit from the abundant data generated in Earth observation campaigns. As such, VQA for remote sensing was identified as one of six promising directions in the agenda of AI for Earth Science Data Analysis~\cite{tuia_toward_2021}.

\begin{figure}[!t]
    \centering
    \includegraphics[width =0.8\textwidth]{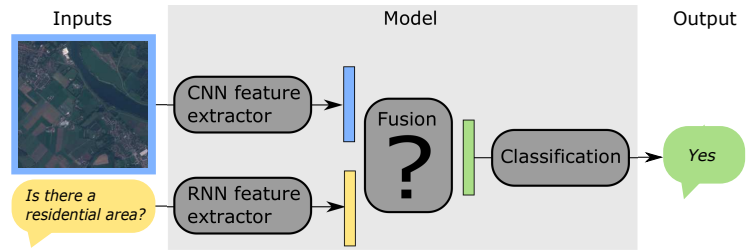}
    \caption{Simplified representation of a VQA model for remote sensing.}
    \label{fig:Model}
\end{figure}

The more often, VQA is based on two data processing streams, one pertaining to image analysis and another focusing on text mining. The feature extractors, generally based on deep learning~\cite{lecun2015deep}, are usually separated and combine their outputs in a dedicated fusion step before predicting an answer category (Fig.~\ref{fig:Model}). This text-image embedding fusion step is central to the task as it not only combines the image and question features, but should also uncover interactions between the two modalities. In its simplest form, the fusion mechanism is an element-wise operation. While straightforward, this type of fusion is limiting in term of interactions, as it requires relevant and matching image and question contents to be aligned at the same index in the latent feature vectors prior to fusion. 
If such ordering cannot be established, a simple fusion operation that restricts the interactions to the same index might not suffice. Therefore, more complex approaches have been proposed~\cite{zhang_multimodal_2020}. While promising in standard VQA benchmarks, these fusion strategies have never been evaluated on remote sensing imagery and their variability. In this paper, we evaluate the effectiveness of these more elaborate fusion strategies when exposed to the variability of remote sensing images. To this end, we build on the RSVQA model and datasets from~\cite{lobry_rsvqa_2020} and compare three different fusion strategies.

\section{Related work}
\label{Related_work}

\paragraph{Visual question answering.}
The task of free-form and open-ended VQA has first been proposed by~\cite{antol_vqa_2015}. The goal is to answer questions about images, where both questions and answers are formulated in natural language. 
The original model of~\cite{antol_vqa_2015} is composed of two parallel feature extractors followed linearly by a fusion and classification. Follow-up works attempt to condition the model on relevant image and question parts, often by employing attention mechanisms~\cite{chen_abc-cnn_2016,yang_stacked_2016}. A different directive focuses on the integration of external knowledge~\cite{wu_ask_2016,wang_explicit_2017}. Finally, compositional models have been suggested to translate the question into a series of simpler processing tasks to be solved in a logical sequence~\cite{andreas_learning_2016,andreas_neural_2016}.

\paragraph{VQA for remote sensing.} %

The first application of VQA to remote sensing images was introduced by~\cite{lobry_rsvqa_2020}. The architecture proposed is in line with~\cite{antol_vqa_2015} and specifically consists of an element-wise multiplication fusion to combine textual and visual features. The authors further published two datasets  (``RSVQA''\footnote[1]{The dataset and the models used in~\cite{lobry_rsvqa_2020} are available at \href{https://rsvqa.sylvainlobry.com}{rsvqa.sylvainlobry.com}.}), consisting of low- (Sentinel-2) and very-high-resolution (aerial) imagery, paired with questions and answers on various tasks (presence/absence classification, object counting, etc.) obtained from OpenStreetMap vector layers.

Recently, three other remote sensing datasets involving a VQA component have been introduced: FloodNet~\cite{rahnemoonfar_floodnet_2020} focuses on natural disasters and contains images acquired with an unmanned aerial vehicle (UAV) after the passage of hurricane Harvey. RSIVQA~\cite{zheng_mutual_2021} is composed of existing classification and object detection datasets. While a few images are annotated by humans for questions/answers, most of the samples had automatically been generated as in the RSVQA dataset. The authors also propose an architecture using a mutual attention and bilinear feature fusion. Finally, RSVQAxBEN~\cite{lobry_RSVQA_BEN_21} is a large-scale (14+ millions image/question/answer triplets) VQA-tailored derivation of the BigEarthNet classification dataset~\cite{BigEarthNet}.

\paragraph{Fusion methods.} 
The fusion step in a conventional VQA model combines both image and question modalities and must therefore encode co-dependencies as efficiently as possible. In theory, an outer product between the feature vectors emerging from the image- and question-specific feature extractors would provide a maximum in terms of feature sharing. However, this quickly becomes intractable, as the number of learnable parameters increases drastically in the following fully-connected layers. Hence, a common compromise is to employ a straightforward element-wise operation (as implemented in~\cite{lobry_rsvqa_2020}), or a concatenation. 
While computationally simple, these approaches rely on the feature extractors and classifier to establish feature correspondences in the latent space during end-to-end training of the model. 
More complex methods have been proposed to enable a richer interaction between elements of both modalities while limiting the number of parameters to learn. 
The first approach, multimodal compact bilinear pooling (MCB~\cite{fukui_multimodal_2016}), uses a random projection on the image and question features before combining them. As such, it consider only a (random) subset of the complete combinatorial between the two modalities. 
Later on,~\cite{kim_hadamard_2017} suggested a low-rank bilinear model (MLB) where the projection of the modalities is learned, fused with Hadamard product and its output projected to the prediction space. 
A factorized bilinear pooling (MFB) was presented by~\cite{yu_multi-modal_2017}, building on MLB by adding a sum pooling to condense the output. An extension introduced by~\cite{yu_beyond_2018} is multimodal factorized high-order pooling (MFH) that applies several MFB to generalize to higher-order pooling (i.e., fusing more than two modalities).
Finally,~\cite{ben_younes_mutan_2017} introduced MUTAN, which employs a matrix decomposition (Tucker decomposition) to break down the full fusion into projections specific to each modality and a smaller learned fusion operation. 
The idea of tensor decomposition was further researched and improved in~\cite{ben_younes_block_2019}, where authors used a block-term decomposition.

\section{Methods}

Our objective is to assess the required level of complexity at the fusion step to express the interplay between features extracted from the question $\mathbf{q}$ and the features extracted from the image $\mathbf{v}$ in VQA for remote sensing. Both vectors $\mathbf{q}$ and $\mathbf{v}$, as well as the resulting vector of the fusion $\mathbf{f}$, are of dimension $n$.
In this work, we employ MCB and MUTAN as representatives for increasing fusion complexity and compare them with the element-wise multiplication of~\cite{lobry_rsvqa_2020}.

The element-wise multiplication ($f_i = v_i \times q_i, \forall i$) is computationally efficient ($\theta(n)$ complexity), but limits the interaction between elements of the textual and visual feature vector, as the $i$th element of the visual vector can only interact with the $i$th element of the textual vector. Moreover, the interaction is unweighted, since there are no additional learnable parameter in the fusion layer.

The MCB fusion strategy uses the Count Sketch algorithm~\cite{charikar_finding_2004}, a dimensionality reduction technique proposed to estimate the frequencies of items in data streams, to avoid computing the outer product of $\mathbf{v}$ and $\mathbf{q}$. In the fusion step of VQA, a Count Sketch projection is applied to each modality independently. Practically, given a vector $\mathbf{x}$ of dimension $n$ and its projection $\mathbf{y}$ of dimension $d$ ($d > n$), two new vectors of dimension $n$ are created: $\mathbf{s}~\in\{-1,1\}^n$ and $\mathbf{h}~\in\llbracket 1, d \rrbracket^n$. Each element of $\mathbf{x}$ is assigned with a sign value from $\mathbf{s}$ and a position in the projected vector $\mathbf{y}$ from $\mathbf{h}$ (and the other $d - n$ elements of $\mathbf{y}$ are set to 0). Randomly sampled from a uniform distribution, $\mathbf{s}$ and $\mathbf{h}$ are fixed at initialization and remain unchanged during the training. Once projected with Count Sketch, the two resulting vectors of $\mathbf{v}$ and $\mathbf{q}$ are combined with a convolution (for efficiency, MCB performs an element-wise product in the frequency domain to this end). Although this approach enables more interactions between the feature vectors, they are random, fixed and determined by the Count Sketch projection and its vectors $\mathbf{s}$ and $\mathbf{h}$. Thus, technically, the fusion itself is not learnt. Moreover, the optimal size of the output dimension $d$ is problem-specific and typically rather large, leading to a large input to the fully connected layer between fusion and classification parts.

\begin{figure}[h]
    \centering
    \includegraphics[width =\textwidth]{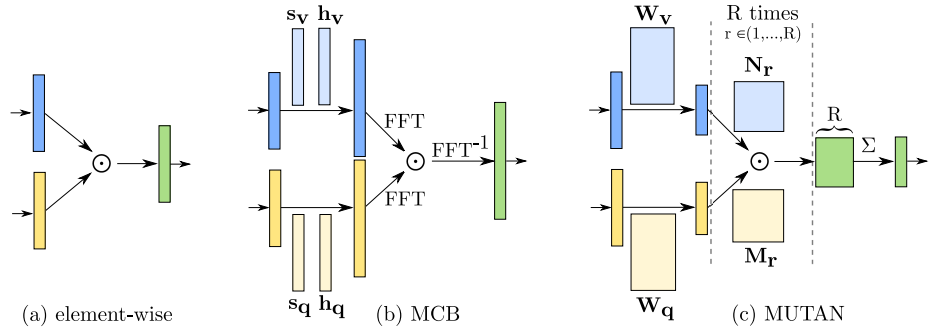}
    \caption{Comparison of the three fusion strategies considered in the paper.}
    \label{fig:Fusion}
\end{figure}

The last method, MUTAN, considers the fully-parameterized 3D tensor operator of a bilinear model and performs a Tucker decomposition~\cite{tucker_mathematical_1966} (described in~\cite{kolda_tensor_2009} as a ``higher-order form of principal component analysis'') to reduce its size. The 3D tensor operator $\mathbf{T}$ is factorized into 3 matrices $\mathbf{W_q}$, $\mathbf{W_v}$, and $\mathbf{W_o}$ and one smaller core tensor $\mathbf{T_c}$ (Figure~\ref{fig:Fusion}). With this decomposition, each factor matrix is given a distinct role in the fusion and can regulate the complexity for its specific modality. The Tucker fusion can be written as: 
$$\mathbf{y} = (( \mathbf{T_c} \times (\mathbf{q} ^T \mathbf{W_q})) \times (\mathbf{v}^T \mathbf{W_v})) \times \mathbf{W_o},$$
where $\mathbf{y}$ is the output vector and the core tensor $\mathbf{T_c}$ is further controlled with a structured sparsity constraint in MUTAN. Two operations are performed first: $\tilde\mathbf{q} = \tanh(\mathbf{q}^T \mathbf{W_q})$ and $\tilde\mathbf{v} = \tanh(\mathbf{v} ^T \mathbf{W_v})$. By combining them, the latent pair representation can be defined as $\mathbf{z} = (\mathbf{T_c} \times \tilde\mathbf{q}) \times \tilde\mathbf{v}$, while the output projected in the prediction space is: $\mathbf{y} = \mathbf{z} ^T \mathbf{W_o}$. With the structured sparsity constraint, $\mathbf{z}$ results from the sum of R multiplications between the question and visual parts: $\mathbf{z} = \sum_{r=1}^{R} (\tilde\mathbf{q} ^T \mathbf{M_r})*(\tilde\mathbf{v} ^T \mathbf{N_r})$, where R is a hyperparameter.

These mechanisms cover different fusion strategies, including a straight-forward combination (point-wise multiplication), one ruled by randomness (MCB) and finally a process fully learnt during training and enabling more interactions as well (MUTAN). Figure~\ref{fig:Fusion} illustrates the three types of fusion we experiment with.

\section{Experiments}
\label{Experiments}
\paragraph{Datasets.}
We evaluate the three fusion strategies on the RSVQA datasets~\cite{lobry_rsvqa_2020}: a low resolution dataset with Sentinel-2 images over the Netherlands (77,232 questions-answers-images triplets) and a high resolution dataset with aerial imagery from the USGS collection (1,066,316 triplets). For the latter, locations cover different regions in the North East Coast of the USA. Training and validation sets contain images of New York City (NY), Long Island (NY) and Portland (ME), while two test sets are provided, one over New York City and Long Island, and a second one over Philadelphia (PA), respectively referred to as NYC and PHL in this paper. The resolutions of images in the low and high resolution datasets are 10m and 15cm, respectively.

\paragraph{Model architecture.}
As depicted in Fig.~\ref{fig:Model}, the fusion operation is preceded by two feature extractors and followed by a classification network. This structure remains unchanged while the fusion operation is investigated. To extract features from the image, we use a ResNet-152~\cite{he2016deep}, pre-trained on ImageNet~\cite{deng_imagenet_2009} and keep all but the ultimate classification layer. For the question part, we use skip-thoughts~\cite{kiros_skip-thought_2015}, pre-trained on BookCorpus~\cite{zhu_aligning_2015}. Both pathways result in a 2048- and 2400- dimensional latent feature vector respectively, reduced to 1200 with fully-connected layers prior to fusion. To predict the answer, the classification is done with one fully-connected layer and one output layer that contains as many classes as there are answers, as in~\cite{lobry_rsvqa_2020}.

\paragraph{Experimental setup.}
In total, we evaluate six models; a baseline with an element-wise multiplication fusion, a model with MCB fusion and a model with MUTAN fusion, one for the low and high resolution each. Each experiment is run three times to compute a mean performance with standard deviation. The performance matrix consists of an accuracy measure for each question type, as well as an average and overall accuracy. All models are trained with the Adam optimizer~\cite{kingma_adam_2017}, a learning rate of $10^{-5}$, and a batch size of 70. The models on low resolution are trained for 150 epochs while those on high resolution are trained for 35 epochs. Finally, we perform an ablation study on the model with MCB fusion on the low resolution dataset portion to test the sensibility of the output dimension of the fusion on the performances: we train and compare models with feature dimensions d = $\{1200, 4000, 8000, 16000, 32000\}$.

\section{Results and discussion}

\paragraph{Low resolution dataset.}
The results for the models on low resolution are displayed in Table~\ref{tab:resultsLR}. A more complex fusion helps the performances of the model in low resolution. However, the improvement varies depending on the question type, with the ``comparison'' questions improving the most, followed by ``counting'' and ``presence'', but to a lesser extent. We observe a strong variability for the accuracy on the question type ``rural/urban''. This is due to the low number of samples for this question type (only one question per image, 1\% of the samples) as well as the subjective choice to differentiate rural and urban (a threshold on the number of buildings was used in~\cite{lobry_rsvqa_2020} to define the image as rural or urban). Overall, MCB performs slightly better than MUTAN, although with a lower performance gap than between MCB and the baseline. However, the difference in the number of parameters to train is considerable. The output dimension of the fusion is 8,000 for MCB and 360 for MUTAN, and the question remains whether the better performance with MCB is the result of the more expressive fusion itself or of the larger capacity in the fully-connected layer used for classification. 

\begin{table}
\caption{Test results on low resolution images, average performance reported with standard deviation in brackets.}\label{tab:resultsLR}
\centering
\begin{tabular}{|l|c|c|c|}
\hline
Question type &  Baseline & MCB & MUTAN\\
\hline
\# of parameters learned & $5.7\times 10^6$ & $7.5\times 10^6$ & $4.4\times 10^6$ \\
\hline
Comparison & 84.44 (0.09) & \textbf{88.22 (0.14)} & 87.52 (0.01) \\
Counting &  68.00 (0.65) & \textbf{70.18 (0.46)} & 69.06 (0.45) \\
Presence & 88.49 (0.25) & \textbf{90.34 (0.47)} & 90.07 (0.21) \\
Rural/urban & \textbf{90.67 (0.47)} & 90.00 (0.82) & 87.67 (2.36)\\
\hline
Average accuracy & 82.90 (0.24) & \textbf{84.69 (0.33)} & 83.58 (0.71) \\
Overall accuracy & 80.86 (0.24) & \textbf{83.55 (0.20)} & 82.84 (0.21) \\
\hline
\end{tabular}
\end{table}

\begin{figure}[t]
    \centering
    \includegraphics[trim=10 130 10 70, clip, width =\textwidth]{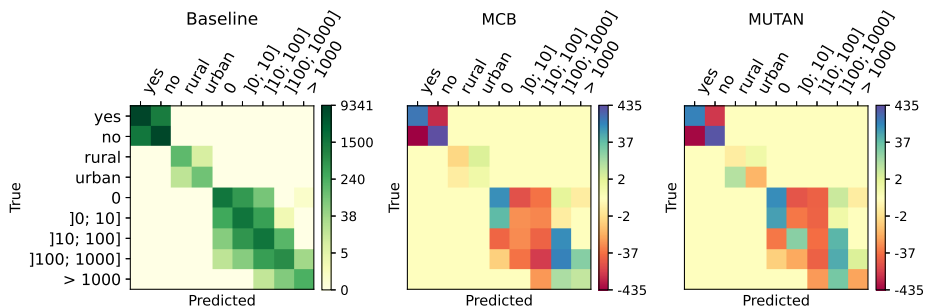}
    \caption{Confusion matrix for the low resolution baseline and differences between the confusion matrix of MCB and MUTAN with the baseline. On the 2nd and 3rd figures, a positive value indicates more predictions with the model compared to the baseline, respectively negative indicates less.}
    \label{fig:confMat_LR}
\end{figure}

The confusion matrix for the baseline and its differences to MCB and MUTAN (Fig.~\ref{fig:confMat_LR}) highlight the capacity of the model to respond with the correct type of answer to each question. The answers ``yes/no'' see most improvements, with the two more elaborated fusion improving the results. Yet, these answers are also the most frequent in the dataset, so it might be reasonable to consider techniques to account for less represented types of answers. Interestingly, MUTAN does slightly worse on the ``rural/urban'' questions, especially ``urban'', and MCB declines on ``rural'' but improves on ``urban''. For ``counting'' questions, the pattern is similar between MCB and MUTAN but we can observe a few variations, the largest class ``more than 1000'' for example is slightly better with MCB, and worse with MUTAN.

As a final result for the low resolution dataset, Table~\ref{tab:ablation_MCB_LR} lists the accuracies obtained with MCB and different output vector dimensions $d$. While the number of parameters increase substantially, the difference in both average and overall accuracy is narrow. The best results are obtained with an output dimension of 8,000 for both overall (OA) and average (AA) accuracies.

\begin{table}
\caption{Results (over the validation set of the low resolution dataset) for the ablation study of the fusion output dimension $d$ in MCB.}\label{tab:ablation_MCB_LR}
\centering
\setlength{\tabcolsep}{5pt}
\begin{tabular}{|r|c|c|c|c|c|}
\hline
$d$ &  1,200 & 4,000 & 8,000 & 16,000 & 32,000\\
\hline
\# of parameters learned ($\times 10^6$) &  $5.7$ & $6.4$ & $7.5$ & $9.5$ & $13.6$\\
\hline
AA & 84.21 & 84.57 & \textbf{85.16} & 85.11 & 84.33 \\
OA & 85.13 & 85.92 & \textbf{86.36} & 86.02 & 86.28 \\
\hline
\end{tabular}
\end{table}

\paragraph{High resolution dataset.}
The results of the models on the two high resolution test sets, NYC and PHL, are displayed in Table~\ref{tab:resultsHR}. An improvement of performance is observed again between the baseline and MCB/MUTAN fusions, but with lower overall improvements compared against the baseline. Again, the largest gain is observed for comparison questions, while the other type of questions also improve, but less. Contrary to the low resolution dataset where MCB was performing better, the difference here is small and included in the standard deviation calculated over the three runs. In line with~\cite{lobry_rsvqa_2020}, a significant drop in performance can be observed in the PHL test set. This comes from the domain shift introduced in this test set, made of images over a city unseen during training. When trying to generalize with this additional PHL test set, the comparison of fusion strategies is consistent but the fusion is impacted as the performance differences are smaller and the variability higher. The domain shift primarily affects the image feature extractor and thus is not really compensated for in the fusion step. Hence, other modelling strategies or larger-scale datasets are needed for better generalization.

\begin{table}
\caption{Test results on high resolution images, average performance reported with standard deviation in brackets.}\label{tab:resultsHR}
\centering
\begin{tabular}{|l|l|c|c|c|c|c|c|}
\hline
Test set & Question type & Baseline & MCB & MUTAN\\
\hline
& \# of parameters learned & $5.7\times 10^6$ & $7.4\times 10^6$ & $4.3\times 10^6$ \\
\hline
\multirow{6}{*}{NYC} & Area & 85.20 (0.14) & 85.77 (0.07) & \textbf{85.80 (0.07)} \\  \hhline{~----}
& Comparison & 88.17 (0.06) & \textbf{90.03 (0.09)} & 90.00 (0.05) \\ \hhline{~----}
& Counting & 68.63 (0.05) & 69.22 (0.04) & \textbf{69.27 (0.05)} \\ \hhline{~----}
& Presence & 90.47 (0.03) & 91.31 (0.10) & \textbf{91.46 (0.02)} \\ \hhline{~----} \hhline{~----}
& Average accuracy & 83.12 (0.01) & 84.08 (0.03) & \textbf{84.13 (0.02)} \\
\hhline{~----}
& Overall accuracy & 83.23 (0.01) & 84.30 (0.02) & \textbf{84.35 (0.02)} \\
\hline\hline
\multirow{6}{*}{PHL} & Area & 75.01 (1.46) & \textbf{75.17 (0.40)} & 74.61 (0.88) \\  \hhline{~----}
& Comparison &  86.20 (0.32) & 87.38 (0.09) & \textbf{87.41 (0.38)} \\ \hhline{~----}
& Counting & 61.47 (0.12) & 61.79 (0.13) & \textbf{61.99 (0.04)} \\ \hhline{~----}
& Presence & 86.36 (0.56) & \textbf{86.82 (0.09)} & 86.72 (0.20) \\ \hhline{~----}
\hhline{~----}
& Average accuracy & 77.26 (0.60) & \textbf{77.79 (0.10)} & 77.68 (0.29) \\  \hhline{~----}
& Overall accuracy & 78.14 (0.48) & \textbf{78.77 (0.06)} & 78.72 (0.24) \\
\hline
\end{tabular}
\end{table}

\begin{figure}[h]
    \centering
    \includegraphics[trim=10 110 10 30, clip, width =\textwidth]{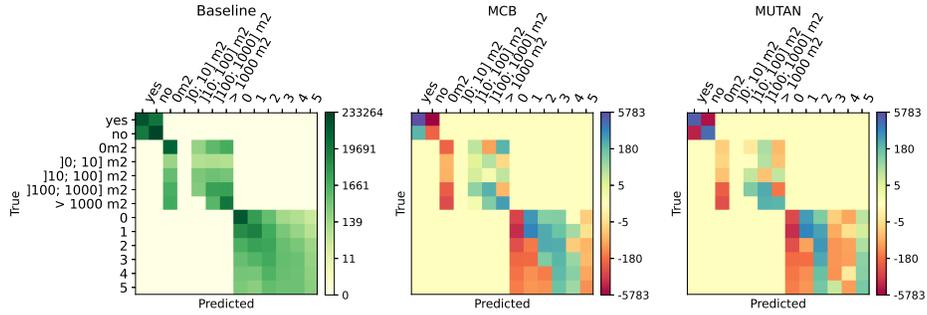}
    \caption{Confusion matrix for the high resolution baseline, along with the differences with MCB and MUTAN (test set NYC, and only the first main classes are represented)}
    \label{fig:confMat_HR}
\end{figure}

The confusion matrices for the high resolution results are displayed in Fig.~\ref{fig:confMat_HR}. For ``yes/no'' answers, MUTAN shows a more consistent improvement compared to MCB. Although both MCB and MUTAN learn to predict wrong small counting values less often, the former does slightly better for counting questions with a few more positive values on the diagonal. The difficulty for the models to answer counting questions is clearly illustrated in these figures.

\section{Conclusion}
We considered a model for visual question answering (VQA) in remote sensing imagery where high-level representations acquired from an image and a question are fused together before predicting an answer. We particularly focused on the fusion component of deep VQA models, which is expected to create meaningful interactions between both modalities, uncovering the key relation to derive a correct answer while maintaining a tractable model. From experiments on the RSVQA dataset, we conclude that a richer fusion is beneficial to the task. We found performances to improve with more elaborate fusion strategies. While MCB shows better accuracy for the low resolution models, the difference is very small for the high resolution models. MUTAN achieves competing results with less than half the number of parameters and therefore constitutes a valuable choice over simpler strategies without compromising computational cost.

\subsubsection{Acknowledgements.}
This work is supported by the European Space Agency through the Open Space Innovation Program, and is part of the project ``An AI assistant to interact with remote sensing images'' led in partnership between the ECEO lab (EPFL) and the $\mathrm{\Phi}$-lab (ESA).

\bibliographystyle{splncs04}
\bibliography{MACLEAN21}

\end{document}